\documentclass[10pt,twocolumn,letterpaper]{article}

\usepackage{cvpr}              

\usepackage{graphicx}
\usepackage{amsmath}
\usepackage{amssymb}
\usepackage{booktabs}

\usepackage{color}
\usepackage{graphicx}
\usepackage{amsmath}
\usepackage{amssymb}
\usepackage{booktabs}
\usepackage{multicol}
\usepackage{multirow}
\usepackage{epsfig}
\usepackage{booktabs}
\usepackage{xfrac}
\usepackage{floatrow}
\usepackage{bbding}
\usepackage[pagebackref,breaklinks,colorlinks]{hyperref}

\usepackage[capitalize]{cleveref}
\crefname{section}{Sec.}{Secs.}
\Crefname{section}{Section}{Sections}
\Crefname{table}{Table}{Tables}
\crefname{table}{Tab.}{Tabs.}

\def\cvprPaperID{*****} 
\def\confName{CVPR}
\def\confYear{2022}

\begin{document}

\title{Dynamic Proposals for Efficient Object Detection}

\author{Yiming Cui\\
University of Florida\\
{\tt\small cuiyiming@ufl.edu}
\and
Linjie Yang\\
ByteDance Inc.\\
{\tt\small yljatthu@gmail.com}
\and 
Ding Liu \\
ByteDance Inc.\\
{\tt\small liudingdavy@gmail.com}
}
\maketitle

\begin{abstract}
   Object detection is a basic computer vision task to localize and categorize objects in a given image. Most state-of-the-art detection methods utilize a fixed number of proposals as an intermediate representation of object candidates, 
which is unable to  adapt to different computational constraints during inference.
In this paper, we propose a simple yet effective method which is adaptive to different computational resources by generating dynamic proposals for object detection. We first design a module to make a single query-based model to be able to inference with different numbers of proposals. Further, we extend it to a dynamic model to choose the number of proposals according to the input image, greatly reducing computational costs.
Our method achieves significant speed-up across a wide range of detection models including two-stage and query-based models while obtaining similar or even better accuracy.
\end{abstract}

\section{Introduction}
Object detection is a fundamental but challenging computer vision task. Given an input image, an algorithm aims to localize and categorize objects in an image simultaneously \cite{detr,liu2020video, zhu2021deformable,Cui_2022_WACV, Ren_2017,bochkovskiy2020yolov4,cui2021tf,liu2020large, redmon2016you,girshick2015fast,law2018cornernet,tian2019fcos,tian2021fcos}. 
\textcolor{black}{To achieve a good performance of object detection, two-stage methods \cite{gao2021fast,ren2015faster,Cai_2019,He_2017} first generate a fixed number of coarse proposals and then refine them to output fine-grained predictions. As one line of successful two-stage methods, R-CNN family \cite{ren2015faster,Cai_2019,He_2017,Cai_2019} utilize region proposal network (RPN) to localize objects roughly and then regions-of-interest features are extracted to output refined predictions. }
\begin{figure} [!tb]
    \centering
    \includegraphics[width=8cm]{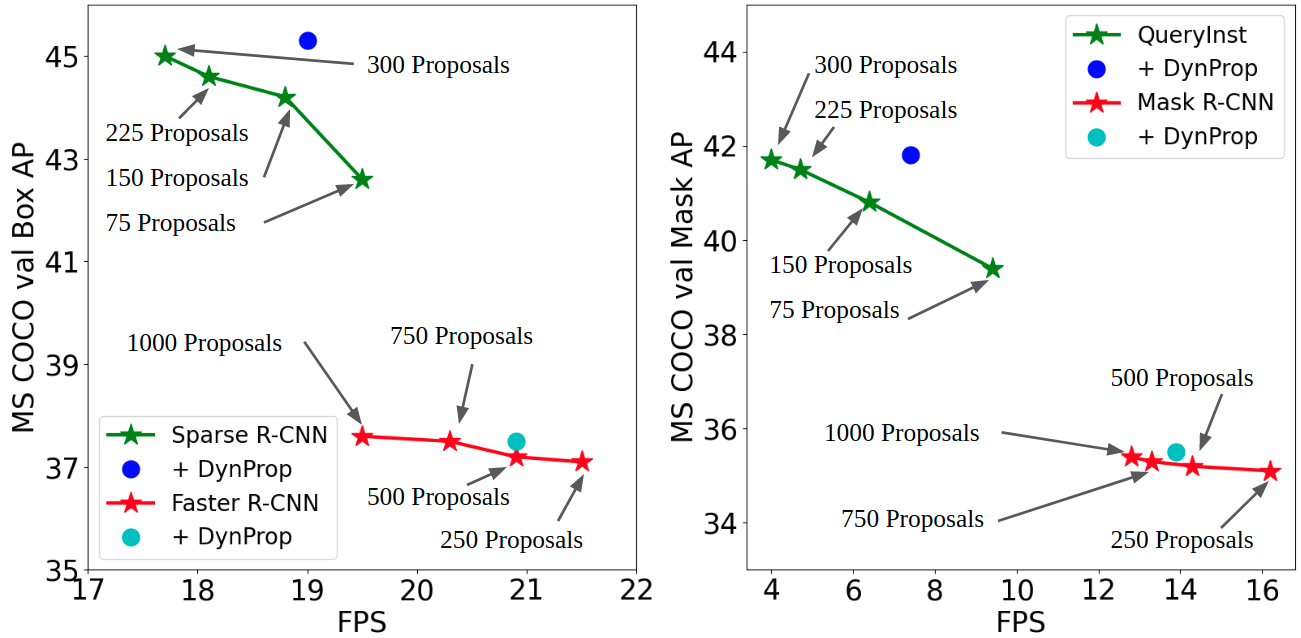}
    \caption{AP $vs.$ FPS on MS COCO \texttt{val} \cite{lin2015microsoft} benchmarks. Integrated with our dynamic proposals, the inference speeds of the four shown detection methods increase by a large margin while maintaining competitive performance. The inference speed is measured with a single TITAN RTX GPU.}
    \label{fig:imageDensity}
\end{figure}
To simplify the procedure of object detection, query-based methods are proposed to remove the manually designed anchor boxes~\cite{detr,zhu2021deformable,sun2021sparse}. Among them, DETR \cite{detr} is a pioneer work to treat object detection as a  direct set prediction problem with multi-stage transformers and learned object queries. 
Sparse R-CNN~\cite{sun2021sparse} designs a query-based set prediction framework based on R-CNN detectors \cite{girshick2015fast,Ren_2017}. By replacing handcrafted proposals with a fixed number of learnable proposals, Sparse R-CNN effectively reduces the number of proposals and avoids many-to-one label assignments. 

Current \textcolor{black}{two-stage and} query-based object detection methods always have a fixed number of proposals regardless of the image contents or computational resources. It is unnecessary to use too many proposals for Sparse R-CNN \cite{sun2021sparse} to detect objects in simple images with only a few objects, since the model with a small number of proposals can already produce a good performance. Also, since the number of proposals is fixed, these models are not re-configurable to strike an accuracy-efficiency balance across devices with various computational budgets. For instance, given a trained model with $300$ proposals, the current methods cannot be directly modified to a model with $100$ proposals during the inference process. 
Therefore, we wonder whether it is possible to use a dynamic number of proposals based on the complexity of the image to improve its efficiency for object detection.

In this paper, we crafted a training strategy to facilitate a single model to adaptively switch the number of proposals according to the hardware constraints. Empirical studies show our model achieves similar performance as individually-trained models under the same number of proposals. Further, we designed a network module to dynamically choose the number of proposals based on the complexity of the input image, which effectively reduces computational costs. Our proposed method is a plug-and-play module which can be easily extended to most of the current two-stage and query-based object detectors. To the best of our knowledge, we are the first to equip object detectors with a dynamic number of proposals to accommodate different computational resources and to improve overall efficiency. Our contributions can be summarized as follows:

\begin{itemize}
\item  We introduce a strategy to train a single object detection model to be adaptive to different computational constraints by switching the number of proposals. 
    An inplace distillation training strategy is added to improve the model variants with fewer proposals by learning from the model with most proposals. 
    \item 
    We design a framework which uses a variable number of proposals for object detection and instance segmentation.
    Our model automatically selects different numbers of proposals based on the complexity of the input image, improving the overall efficiency of the model while still achieving similar accuracy as original static models.
    \item Our proposed approach is a plug-and-play module which can be easily integrated with most of the current two-stage and query-based methods for object detection and instance segmentation. For example, when applying our method on QueryInst~\cite{Fang_2021_ICCV}, the inference speed increases by around $80\%$ with similar or even better precision, as shown in Figure \ref{fig:imageDensity}.
\end{itemize}
\section{Related Works}
\noindent\textbf{Object detection.} Two-stage object detectors such as the R-CNN family \cite{Ren_2017,girshick2015fast,ren2015faster} use a large number of predefined anchor boxes to mark the initial locations of candidate objects. To improve the efficiency and inference speed of object detectors, one-stage methods \cite{redmon2016you,bochkovskiy2020yolov4} are introduced which tackle the object detection tasks without using region proposals. Recently, query-based methods \cite{detr,zhu2021deformable,sun2021sparse} replaced anchor boxes and region proposals with learned proposals / queries. DETR \cite{detr} modifies an encoder-decoder architecture based on transformers \cite{vaswani2017attention} to generate a sequence of prediction outputs and introduces a set loss function to perform bipartite matching between predicted and ground-truth objects. Deformable-DETR \cite{zhu2021deformable} improves convergence of DETR by improving feature spatial resolutions. 
Sparse R-CNN \cite{sun2021sparse} introduces a fixed sparse set of learned object proposals to classify and localize objects in the image with dynamic heads and generate final predictions directly without post-processing methods like non-maximum suppression. 

\noindent\textbf{Instance segmentation.} As a natural extension of object detection with bounding boxes, instance segmentation requires the algorithm to assign every object with a pixel-level mask \cite{chen2020blendmask, Cao_SipMask_ECCV_2020, liu2021sg, yang2019video}. Following the frameworks of object detection, Mask R-CNN \cite{He_2017} introduces a mask head to Faster R-CNN \cite{Ren_2017} to produce instance-level masks. Similarly, one-stage instance segmentation frameworks such as YOLACT \cite{yolact-iccv2019,yolact-plus-tpami2020}, SipMask \cite{Cao_SipMask_ECCV_2020} and SOLO \cite{wang2020solo,wang2020solov2} are introduced to balance the inference time and accuracy. Recently, QueryInst \cite{Fang_2021_ICCV} extended query-based object detection method Sparse R-CNN \cite{sun2021sparse} to instance segmentation task by adding a dynamic mask head and parallel supervision. However, all the two-stage and query-based methods mentioned above have a fixed number of proposals which is not adaptive to images with different objects as well as devices with various computational resource constraints.

\noindent\textbf{Dynamic networks.} 
Recently, slimmable networks \cite{yu2018slimmable,yu2019universally,yu2019autoslim} were introduced to be adaptive to different devices and achieve even better performance compared with their stand-alone counterparts trained separately. DS-Net \cite{li2021dynamic} designs a dynamic network slimming approach to achieve hardware-aware efficiency by dynamically adjusting the structures of the networks during the inference time based on the inputs, while preserving the parameters of the model in hardware statically and contiguously to avoid the extra cost for model development and deployment. \textcolor{black}{Different from these slimmable networks \cite{yu2018slimmable,yu2019universally,yu2019autoslim} which apply switchable operations channel-wisely, our proposed method makes the number of proposals, a new dimension commonly occurred in object detection and instance segmentation models, become adjustable in inference time. Note that a proposal represents information of an object candidate and is drastically different from a feature channel that does not have explicit semantic meanings.} 

\section{Methodology}
\begin{figure*}[!bt]
    \centering
    \includegraphics[width=0.8\textwidth]{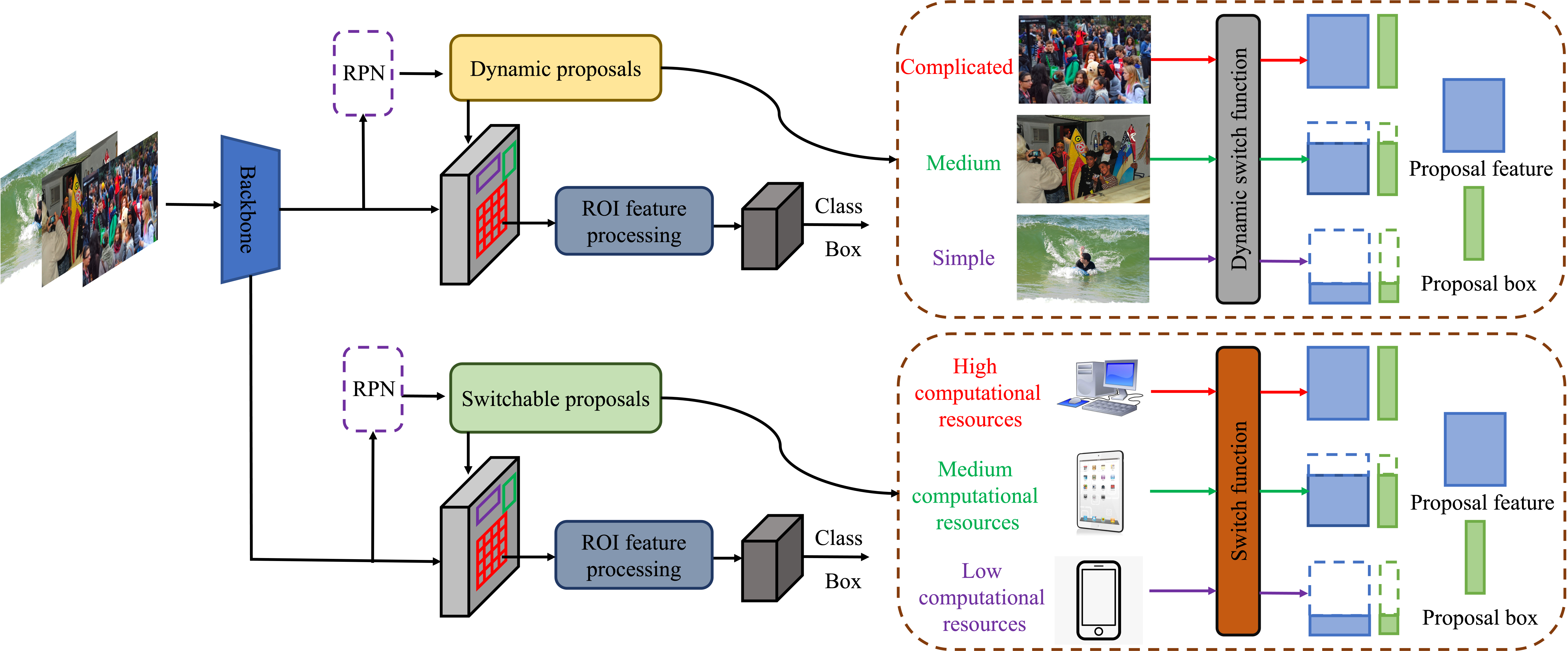}
    \caption{The framework of our proposed method which uses switchable proposals for object detection on devices with different computational resources and dynamic  proposals for efficient object detection. The dotted box of RPN represents the difference of the two model variants: two-stage (with RPN) and query-based (without RPN). For input images with few objects or devices with low computational resources, we use fewer proposals (shown as purple arrows) while for the complicated cases or on high computational resources, we use most proposals (shown as red arrows).}
    \label{fig:compareProposals}
    \vspace{-0.2cm}
\end{figure*}

The key idea of our proposed method is to replace the fixed number of proposals with a dynamic size in the current object detection methods. Instead of using fixed proposals, our model chooses different numbers of proposals based on the content of input images or the current computational resources, as shown in Figure \ref{fig:compareProposals}. Our method can be easily plugged with most two-stage and query-based detection methods.

In the following sections, we first review the current object detection methods with proposals and introduce a training strategy with switchable proposals to make our model adaptive to different configurations during inference. Then, we extend switchable proposals to dynamic proposals so that the proposal numbers can be adjusted based on the input images adaptively. Finally, we 
introduce an inplace distillation strategy to transfer knowledge in our model from the networks with more proposals to those with fewer proposals in each training iteration, which significantly improves the overall performance of our model.
\subsection{Revisiting Two-stage and Query-based Methods}
\textcolor{black}{Both traditional two-stage detection methods and recent query-based models both have proposals as an intermediate representation for object candidates. In two-stage models, coarse proposals are first generated by the RPN module and then refined to output the final predictions. Query-based methods replace the RPN module with learnable queries to generate the object candidates for final prediction. The inference speed and performance of two-stage and query-based methods are heavily influenced by the number of proposals. For example, when the number of proposals decreases from 300 to 75, the inference speed of QueryInst \cite{Fang_2021_ICCV} can increase from $3.7$ FPS to $9.4$ FPS while the mAP drops from $41.7$ to $39.4$.} 

For two-stage detection models, proposals are generated by Region Proposal Networks (RPN)~\cite{ren2015faster}, and are further applied on the feature map to extract corresponding ROI features. Given the feature map $x^\text{FPN}$ from FPN \cite{lin2017feature}, the prediction of two-stage object detectors is generated by:
\begin{equation}
\begin{aligned}
    b &\leftarrow \texttt{RPN}\left(x^{\text{FPN}}\right)\\
    p_{\text{box}} &\leftarrow T\left(x^{\text{FPN}}, b\right) 
\end{aligned}
\end{equation}
\textcolor{black}{where $b$ denotes the proposal outputs from RPN \cite{ren2015faster} and $T$ represents the task networks to generate the classification, bounding box and/or mask predictions $p_{\text{box}}$ for each object proposal. Although number of proposals generated by RPN is determined by number of regions predicated with a sufficient foreground likelihood in the image, the number is often too large and is truncated at a fixed number in main implementations of RPNs~\cite{mmdetection}. For example, Faster R-CNN uses 1000 as the default number of proposals. }

Query-based methods use learned proposals instead of manually defined anchor boxes as the initial candidate locations. Query-based methods are often equipped with multiple refinement stages after initial bounding boxes to refine the location of interested objects. For example, Sparse R-CNN \cite{sun2021sparse} has $6$ refinement stages. 
Given the feature map $x^\text{FPN}$ from FPN \cite{lin2017feature}, the process of one stage can be formalized as:
\begin{equation}
    \begin{aligned}
   \left(b_t, q_t\right) &\leftarrow \mathcal{S}_t\left(x^\text{FPN}, b_{t-1}, q_{t-1} \right), t = 1, 2, \dots, 6
   \end{aligned}
   \label{eq: stageFullProposals}
\end{equation}
where $q_t \in \mathbb{R} ^ {N \times d}$ and $b_t \in \mathbb{R} ^ {N \times 4}$ represent the proposal features and proposal boxes \cite{sun2021sparse,Fang_2021_ICCV}. $N$ and $d$ denote the number and dimension of proposals respectively. For the $t$-th stage $\mathcal{S}_t$, the model takes $b_{t-1}, q_{t-1}$ and $x^\text{FPN}$ as inputs and generate the corresponding proposal boxes $b_t$ and proposal features $q_t$ for the next stage. At the last stage,  proposal features $q_t$ and boxes $b_t$ are fed into task networks to generate bounding boxes for regression and categories for classification. Sparse R-CNN applies set prediction loss which produces an optimal bipartite matching between predictions and ground truth \cite{sun2021sparse,detr,zhu2021deformable} and the loss can be summarize as:
\begin{equation}
     \mathcal{L} = \lambda_{cls} \cdot \mathcal{L}_{cls} + \lambda_{L1} \cdot \mathcal{L}_{L1} + \lambda_{giou} \cdot \mathcal{L}_{giou}
    \label{eq: sparsercnnLoss}
\end{equation}
where $\mathcal{L}_{cls}$ is the loss for category classification and $\mathcal{L}_{L1}, \mathcal{L}_{giou}$ are losses for bounding boxes regression. $\lambda_{cls}, \lambda_{L1}$ and $\lambda_{giou}$ are hyperparameters to weighted average the losses mentioned above. For two-stage and query-based detection methods, the numbers of proposals are always fixed in both training and evaluation, regardless of the complexity of the image. 



\subsection{Switchable Proposals}

To make our model adaptive to different numbers of proposals, instead of using fixed $N$ proposals $q_0 \in \mathbb{R} ^ {N \times d}$, we use a switch function $\mathcal{F}_{s}(N, \delta)$ to determine the number of proposals $N_{s}$ where $\delta$ is a configurable parameter with value range $(0,1]$.
The formulation of $\mathcal{F}_{s}(N, \delta)$ is, 

\begin{equation}
\begin{aligned}
    N_{s} &= \mathcal{F}_{s}(N, \delta) 
=\left\lceil\left\lceil\delta\theta\right\rceil \frac{N}{\theta}\right\rceil\\
     \end{aligned}
    \label{eq: switch}
\end{equation}
where $\left\lceil\right\rceil$ represents the ceiling function and $\theta$ is a fixed positive integer. With different values of $\delta$, $N_{s}$ have $\theta$ discrete choices given by $\frac{1}{\theta}N$, $\frac{2}{\theta}N$, $...$, $N$. Thus $\theta$ is named number of configurations. Given a fixed $N$, we only choose $\theta$ so that $N$ is divisible by $\theta$ to make $N_s$ an integer.

To equip query-based methods with switchable proposals, we slice a subset of the proposal features  $q_0^{s}$ and a subset of the proposal boxes $b_0^{s}$ from the original proposal features $q_0$ and proposal boxes $b_0$ using 

\begin{equation}
\begin{aligned}
q_0^{s} &= \mathcal{G}(q_0, N_s)\\
     b_0^{s} &= \mathcal{G}(b_0, N_s)
     \end{aligned}
     \label{eq: slice}
\end{equation}
where $\mathcal{G}(\cdot, N_{s})$ is a sampling function to select $N_{s}$ proposals from $N$ total proposals, $q_0^{s} \in \mathbb{R} ^ {N_{s} \times d}$, and $b_0^{s} \in \mathbb{R} ^ {N_{s} \times 4}$. Afterwards each stage of the query-based method from Equation~\ref{eq: stageFullProposals} will be updated as: 
\begin{equation}
   \left(b_t^{s}, q_t^{s}\right) \leftarrow \mathcal{S}_t\left(x^\text{FPN}, b_{t-1}^{s}, q_{t-1}^{s} \right), t = 1, 2, \dots, 6
\label{eq: stageFewProposals}
\end{equation}

For two-stage object detection methods\cite{He_2017,Cai_2019,kirillov2020pointrend}, our approach can also be easily plugged in to adjust number of proposals produced by RPNs. In this case, switchable proposals in Equation \ref{eq: slice} and \ref{eq: stageFewProposals} can be extended as: 
\begin{equation}
\begin{aligned}
    b^s &= \mathcal{G}\left(b, N_s\right) \\
    p_{\text{box}} ^ s &\leftarrow T\left(x^{\text{FPN}}, b^s\right)
\end{aligned}    
\end{equation}

Note that a detection model trained with a fixed number of proposals is not suitable to be directly used with different number of proposals in inference. For two-stage models, only choosing the top proposals will reduce the recall rate and degenerate detection precision. For query-based methods, the performance of the models greatly rely on the information propagation across different proposals in the refinement stages. Removing a large number of proposal directly undermines this propagation procedure and leads to significant degeneration. See Sec.~\ref{exp:switchable} for results of this direct approach.

During the training process, we generate $\delta$ with a uniform distribution $\delta \sim U\left(0, 1\right)$. For the sampling function $\mathcal{G}$, we simply choose the first $N_{s}$ elements out of $N$ elements. 
In our approach, models with different numbers of proposals share the same set of parameters and are jointly optimized during the training process.
During inference, the model can be executed with different numbers of $N_s$, with different computational costs and slightly different performance. 
Depending on the computational resource constraints, we can change $\delta$ and $N_{s}$ accordingly so that the computational cost of the model meets the constraints. When there are enough computational resources, $N_{s}= N$ proposals will be used for inference. When the device can only accommodate less computational resources, for example, a laptop reduces the frequency of CPUs due to heat accumulation, our model can simply switch the configurations to reduce $N_{s}$ to $0.5N$ without the need to reload another model. We thus name this method \emph{Switchable Proposals} for its capability to switch among different configurations of proposals.

\subsection{Dynamic Proposals}
The switchable proposals described in the previous subsection facilitate a two-stage or a query-based detection model to be executed under different numbers of proposals. 
In such cases, the number of proposals is chosen according to external resources, rather than the content of the image. To relate the number of proposals and the computational cost with the content of the image, we use the number of objects in the image 
as the guidance to generate dynamic numbers of proposals. During the training process, 
we estimate the number of objects in the image that is denoted as $\tilde{n}$. We then replace the original variable $\delta$ with a deterministic function $\delta(\tilde{n})$ defined as,
\begin{equation}
    \begin{aligned}
 \delta(\tilde{n}) = \min (\frac{\tilde{n}}{K}, 1)
 \end{aligned}
    \label{eq: delta}
\end{equation}
where $K$ is a fixed parameter, chosen to be a typical number of objects in a complex image. Thus $\delta$ linearly grows with the predicted number of objects upper bounded at $1$. The new dynamic number of proposals $N_d$ is given by

\begin{equation}
    \begin{aligned}
    N_{d} &= \mathcal{F}_{d}\left(N, \delta(\tilde{n})\right) 
    =\left\lceil\left\lceil\theta\delta\left(\tilde{n}\right) \right\rceil \frac{N}{\theta}\right\rceil\\ 
     \end{aligned}
    \label{eq: switchGT}
\end{equation}

\begin{table*}[!tb]
    \centering
    \begin{tabular}{c|c|c|c|c|c|c|c|c|c|c}
    \toprule
    \multirow{2}{*}{Model} & \multirow{2}{*}{Method} & \multicolumn{2}{c|}{0.25N} & \multicolumn{2}{c|}{0.5N} & \multicolumn{2}{c|}{0.75N} & \multicolumn{2}{c|}{N} & \multirow{2}{*}{Train Time}\\
    \cline{3-10} 
    \rule{0pt}{10pt}
    & & AP & AR & AP & AR & AP & AR & AP & AR\\
    \midrule
    \multirow{4}{*}{Sparse R-CNN} & Individual & 42.6 & 58.3 & 44.2 & 63.2 & 44.6 & 64.2 & 45.0 & 66.7 & 110 GPU hours\\
    & Switchable & 42.6 & 59.7 & 44.3 & 64.4 & 44.7 & 66.0 & 45.3 & 66.7 & 73 GPU hours\\
    & Naive selection & 26.5 & 47.4 & 37.2 & 58.0 & 43.5 & 64.4 & 45.0 & 66.7 & -\\
    \cline{3-10}
    \rule{0pt}{10pt}
    & FPS & \multicolumn{2}{c|}{19.5} & \multicolumn{2}{c|}{18.8} & \multicolumn{2}{c|}{18.1} & \multicolumn{2}{c|}{17.7} & -\\
    \midrule
    \multirow{4}{*}{QueryInst} & Individual & 39.4 & 53.5 & 40.8 & 57.2 & 41.5 & 58.8 & 41.7 & 59.9 & 200 GPU hours\\
    & Switchable & 39.4 & 53.2 & 40.9 & 57.3 & 41.1 & 59.0 & 41.4 & 60.0 & 98 GPU hours\\
    & Naive selection & 28.2 & 47.5 & 37.7 & 55.6 & 41.0 & 58.9 & 41.7 & 59.9 & -\\
    \cline{3-10} 
    \rule{0pt}{10pt}
    & FPS & \multicolumn{2}{c|}{9.4} & \multicolumn{2}{c|}{6.4} & \multicolumn{2}{c|}{4.7} & \multicolumn{2}{c|}{3.7} & - \\
    \midrule
    \multirow{4}{*}{Mask R-CNN} & Individual & 35.1 & 46.3 & 35.2 & 47.0 & 35.3 & 47.4 & 35.4 & 47.9 & 62 GPU hours\\
    & Switchable &  35.1 & 46.1 & 35.1 & 46.8 & 35.3 & 47.4 & 35.4 & 47.9 & 19 GPU hours \\
    & Naive selection & 34.7 & 46.1 & 35.1 & 46.8 & 35.4 & 47.7 & 35.4 & 48.1 & - \\
    \cline{3-10} 
    \rule{0pt}{10pt}
    & FPS & \multicolumn{2}{c|}{16.2} & \multicolumn{2}{c|}{14.3} & \multicolumn{2}{c|}{13.3} & \multicolumn{2}{c|}{12.8} & -\\
    \bottomrule
    \end{tabular}
    \caption{Comparisons between original methods trained with multiple configurations individually and jointly trained with switchable proposals on MS COCO \texttt{val}.}
    \label{tab: switchable}
\end{table*}
In query-based models, the dynamic proposal feature  $q_0^{d}$ and box $b_0^{d}$ are sliced from the original $q_0$ and $b_0$ using 
\begin{equation}
\begin{aligned}
q_0^{d} &= \mathcal{G}(q_0, N_d)\\
     b_0^{d} &= \mathcal{G}(b_0, N_d)
     \end{aligned}
     \label{eq: slice2}
\end{equation}
In two-stage models, the object proposals are sampled from the original proposals generated by the RPN using the ratio defined in Equation ~\ref{eq: switchGT},
\begin{equation}
\begin{aligned}
    b^d &= \mathcal{G}\left(b_p, N_d\right) \\
    p_{\text{box}} ^ d &\leftarrow T\left(x^{\text{FPN}}, b^d\right)
\end{aligned}    
\end{equation}

Meanwhile, we design a lightweight network $\mathcal{M}$ to estimate the number of objects $\tilde{n}$ in the current image given the feature maps $x^\text{FPN}$. Although only holistic image features are used to predict the number of objects, it is already informative enough to give a reasonable prediction. Mean square error (MSE) between $\tilde{n}$ and ground truth number of objects $n_{g}$ is calculated as the loss $\mathcal{L}_{est}$ to optimize the network $\mathcal{M}$, as Equation \ref{eq: objectEst}. 
\begin{equation}
\begin{aligned}
    \mathcal{L}_{est} &= \mathcal{L}_\text{MSE} \left(\tilde{n}, n_{g}\right) \\
    &= \mathcal{L}_\text{MSE} \left(\mathcal{M}\left(x^\text{FPN}\right), n_{g}\right)
    \label{eq: objectEst}
\end{aligned}
\end{equation}
We use the same sampling function $\mathcal{G}$ as switchable proposals in Equation~\ref{eq: slice2}. In the training process, $\tilde{n}$ is set to be the current estimation of numbers of objects, $\alpha$ and $\beta$ are set to 0 and 1, respectively. During inference, given an input image, we estimate the number of objects $\tilde{n}$ in the image and use $\mathcal{F}_{d}$ to determine the number of proposals for object detection. For images with fewer objects, where $\tilde{n}$ is small, we use fewer proposals for fast inference. For those with more objects and complex structures, $\mathcal{F}_{d}$ will generate a large $N_{d}$ to provide better coverage of proposals on the target objects. 
This scheme is named \emph{Dynamic Proposals} for its capability to dynamically select the number of proposals based on input complexity.
\begin{table*}[!tb]
    \centering
    \begin{tabular}{c|c|c|c|c|c|c|c|c}
    \toprule
        \multicolumn{2}{c|}{Model} & AP & AP(S) & AP(M) & AP(L) & AR & FPS & Speed up \\
    \midrule
    \multirow{9}{*}{Obj. Det.} & RetinaNet \cite{lin2017focal} & 37.4 & 20.0 & 40.7 & 49.7 & 53.9 & 18.1 & -\\
    & DETR \cite{detr} & 40.1 & 18.3 & 43.3 & 59.5 & 56.5 & 23.1 & -\\
    & Deformable-DETR \cite{zhu2021deformable} & 44.5 & 28.0 & 47.8 & 58.8 & 63.4 & 13.6 & -\\
    \cline{2-9}
    \rule{0pt}{10pt} 
    & Faster R-CNN \cite{ren2015faster} & 37.4 & 21.2 & 41.0 & 48.1 & 51.7 & 19.5  & \multirow{2}{*}{7.2$\%$}\\
    & Faster R-CNN + DP & 37.5 & 20.2 & 40.2 & 49.8 & 48.8 & 20.9\\
    \cline{9-9}
    & Cascade R-CNN \cite{Cai_2019} & 40.3 & 22.5 & 43.8 & 52.9 & 54.3 & 16.6 & \multirow{2}{*}{16.3$\%$}\\
    & Cascade R-CNN + DP & 40.0 & 21.0 & 42.9 & 53.5 & 49.1 & 19.3 \\
    \cline{9-9}
    & Sparse R-CNN \cite{sun2021sparse} & 45.0 & 28.0 & 47.6 & 59.5 & 66.7 & 17.7 & \multirow{2}{*}{7.4$\%$}\\
    & Sparse R-CNN + DP & 45.3 & 27.5 & 48.4 & 59.8 & 65.4 & 19.0\\
    \midrule
    \multirow{8}{*}{Ins. Seg.}
    & Mask R-CNN \cite{He_2017}& 35.4 & 16.6 & 38.2 & 52.5 & 48.1 & 12.8 & \multirow{2}{*}{8.6$\%$}\\
    & Mask R-CNN + DP & 35.3 & 16.4 & 37.4 & 53.0 & 46.3  & 13.9\\
        \cline{9-9}
    & Cascade Mask R-CNN \cite{Cai_2019} & 35.9 & 17.3 & 38.0 & 53.1 & 48.3 & 10.5 & \multirow{2}{*}{30.5$\%$}\\
    & Cascade Mask R-CNN + DP & 35.7 & 15.9 & 37.6 & 53.4 & 45.1 & 13.7\\
    \cline{9-9}
    & QueryInst \cite{Fang_2021_ICCV} & 41.7 & 23.2 & 44.7 & 60.5 & 60.6 & 4.0 & \multirow{2}{*}{85.0$\%$}\\
    & QueryInst + DP & 41.7 & 22.6 & 44.8 & 59.9 & 59.6 & 7.4\\
    \cline{9-9}
    & QueryInst$^\dagger$ & 42.9 & 24.5 & 45.7 & 60.8 & 61.1 & 3.6 &
    \multirow{2}{*}{44.5$\%$}\\
    & QueryInst + DP $^\dagger$ & 43.1 & 23.7 & 45.6 & 61.2 & 60.3  & 5.2 \\
    \bottomrule
    \end{tabular}
    \caption{Comparison of two-stage and query-based methods combined with dynamic proposals and the baseline models on object detection and instance segmentation on MS COCO \texttt{val} \cite{lin2015microsoft}. ``DP'' denotes models combined with dynamic proposals. $\dagger$ denotes models with ResNet-101 as backbone.}
    \label{tab: objectDetection}
    \vspace{-0.2cm}
\end{table*}
\subsection{Inplace Distillation}
With the switchable or dynamic proposals, a single model already embedded a sequence of model variants characterized by different numbers of proposals. During training, the full model, i.e. the model variant with $N$ proposals naturally learns a better feature representation. The knowledge of the full model can be leveraged to improve the model variants with fewer proposals. Towards this end, we introduce an inplace distillation strategy inspired by \cite{yu2019autoslim,yu2019universally}.

In our approach, we treat the full model with $N$ proposals as the teacher model and those with fewer proposals as the student models. During training, in addition to the forward procedure in Equation \ref{eq: stageFewProposals} for an arbitrary student model, we also conduct a forward procedure for the teacher model as in Equation \ref{eq: stageFullProposals}. Distillation can be applied to both switchable and dynamic proposals. We take dynamic proposal in query-based methods as an example and introduce the detailed procedure as follows. At each stage $t$, we calculate the Region of Interest (ROI) features $f_{t}$ and $f_{t}^{d}$ given the input ROI boxes $b_t$ and $b_t^{d}$. The distillation loss at stage $t$ is a weight sum of the mean square errors between $f_t, f_t^{d}$ and $q_t, q_t^{d}$ as Equation \ref{eq: roiPool}. The final distillation loss $\mathcal{L}_{dst}$ sums over different stages and is formulated as Equation \ref{eq: distillLoss}.
\begin{equation}
    \begin{aligned}
        f_t &= \texttt{ROIPool}\left(x^\text{FPN}, b_t\right) \\
        f_t^{d} &= \texttt{ROIPool}\left(x^\text{FPN}, b_t^{d}\right)\\
        \label{eq: roiPool}
    \end{aligned}
\end{equation}
\begin{equation}
    \begin{aligned}
        \mathcal{L}_{dst} &= \sum_t\big( c_1 \mathcal{L}^t_\text{MSE}\left(\mathcal{G}(f_t, N_d),f_t^{d}\right) \\
        &+ c_2 \mathcal{L}^t_\text{MSE}\left(\mathcal{G}(q_t, N_d), q_t^{d}\right)\big) 
    \end{aligned}
    \label{eq: distillLoss}
\end{equation}
where $c_1$ and $c_2$ are two coefficients. Note that we only apply the distillation loss on the feature maps of each object proposal between the teacher and student models. No distillation loss is applied to the output class predictions and bounding box predictions mainly because even powerful detection models produce a considerable number of incorrect predictions. We deliberately make the student model to mimic features learned by the teacher, rather than the outputs.

Note that we only apply inplace distillation to query-base methods rather than two-stage models, since two-stage models use the same ROI features for the student model and the teacher model, making it unnecessary to conduct the distillation step. 


\section{Experiments}
\subsection{Implementation Details}
\noindent\textbf{Experiments setup.}  We conduct experiments on the challenging MS COCO dataset \cite{lin2015microsoft} and Cityscapes \cite{Cordts2016Cityscapes}
following \cite{mmdetection,Fang_2021_ICCV,sun2021sparse}  \textcolor{black}{where the number of objects varies quite much}. \textcolor{black}{In MS COCO, there are 7 objects on average, 1 object at least and 93 objects at most in one image while in Cityscapes dataset there are 21 objects on average, 1 object at least and 120 objects at most in one image.} 
The default backbone is ResNet-50 \cite{he2015deep} with FPN \cite{lin2017feature} unless otherwise specified. On other experimental details like pipelines for data augmentation, training and inference setups, we follow the corresponding papers for a fair comparison. 
During training, we use $8$ Tesla A100 GPUs. During inference, all the methods are evaluated on a TITAN RTX GPU.
On Cityscapes, we follow the pipeline in QueryInst \cite{Fang_2021_ICCV} to finetune the weights pretrained on the MS COCO train2017 split on Cityscapes, and use the same training and inference setups as those in MS COCO experiments.

\noindent\textbf{Implementation.} 
For the object number estimation network $\mathcal{M}$, a global max pooling layer is deployed to reduce the spatial dimension of the feature map,  whose outputs are fed into a two-layer fully connected layer to predict the numbers of objects. A ReLU layer is applied to clamp negative outputs since the numbers of objects should be non-negative.
As default, we choose $\alpha = 0, \beta = 1, \theta = 4$ for all the experiments unless otherwise stated. On experiments with two-stage models such as Cascade R-CNN and Mask R-CNN, $N$ is set to $1000$, same as the original number of proposals in their default settings. On experiments with query-based models (Sparse R-CNN and QueryInst), we set $N = 300$, $c_1 = 0.1$, and $c_2 = 1$. $K$ is set to $20$ on MS COCO and $75$ on Cityscapes since Cityscapes has more objects in each image on average.

\subsection{Main Results}
In this section, we conduct experiments on switchable proposals and dynamic proposals with two-stage detection models (Cascade R-CNN, Mask R-CNN etc) and query-based models (Sparse R-CNN and QueryInst) on object detection and instance segmentation tasks.
\begin{table*}[!tb]
    \centering
    \begin{tabular}{c|c|c|c|c|c|c|c}
    \toprule
    Object Detection & AP & AP$_\text{50}$ & FPS & Instance Segmentation & AP & AP$_\text{50}$ & FPS  \\
    \midrule
    Faster R-CNN  \cite{Ren_2017} & 40.3 & 65.3 & 11.3 & Mask R-CNN \cite{He_2017} & 36.4 & 61.8 & 3.5\\
    Faster R-CNN + DP & 40.0 & 65.4 & 13.4 & Mask R-CNN + DP & 36.1 & 61.2 & 5.1\\
    Sparse R-CNN \cite{sun2021sparse} & 41.8 & 65.6 & 10.7 & QueryInst \cite{Fang_2021_ICCV} & 39.0 & 64.3 & 0.5\\
    Sparse R-CNN + DP & 41.1 & 66.4 & 11.8 & QueryInst + DP & 38.2 & 62.8 & 1.1\\
     \bottomrule
    \end{tabular}
    \caption{Comparison with the baseline models on object detection and instance segmentation on Cityscapes \texttt{val}. ``DP'' denotes models combined with dynamic proposals.}
    \label{tab: cityscapes}
    \vspace{-0.2cm}
\end{table*}

\noindent\textbf{Switchable Proposals}
\label{exp:switchable}
We first conduct experiments on switchable proposals on object detection with both two-stage and query-based models, including Mask R-CNN~\cite{He_2017}, Sparse R-CNN~\cite{sun2021sparse}, and QueryInst~\cite{Fang_2021_ICCV}.
We train $4$ baseline models with different numbers of proposals, namely $0.25N, 0.5N, 0.75N$ and $N$, using the original model settings individually and use their corresponding number of proposals during the inference process, denoted as ``Individual". Meanwhile, we use $N$ total proposals and Equation \ref{eq: switch} as the way to assign $N_{s}$ during the training process. During the inference process, we assign $N_{s}$ with the four ratios to get four inference results, which is denoted as ``Switchable". Note the switchable results are obtained from a single model with switchable proposals. Finally, we select the baseline models with $N$ proposals and naively select the first $0.25N$, $0.5N$, and $0.75N$ elements from the proposals during inference to produce $4$ inference results, denoted as ``Naive Selection". Table \ref{tab: switchable} summarizes the comparison among these trained models with different numbers of proposals during inference processes. 

From Table \ref{tab: switchable}, the performances on AP and AR of separately and jointly training are quite similar given the same number of proposals. The FPS of Switchable and Individual models under the same number of proposals are the same since they have exactly the same computation procedure. Our Switchable model only needs about half total training time of the Individual approach, which needs to train $4$ models separately. For the Naive Selection
setup, the performance drops significantly on Sparse R-CNN and QueryInst compared with Switchable models. 
The two-stage model Mask R-CNN suffers less from the Naive Selection, but is still worse than its Switchable counterpart by more than $0.5\%$ in AP on the variant with $0.25N$ proposals.
This experiment shows that by naively selecting the first $N_{s}$ proposals of the original detection methods, the performance degenerates significantly. Therefore, the original models are not suitable to be directly used for different numbers of proposals by simply sampling the original proposals during the inference process. However, by using our switchable proposals to train the two-stage and query-based methods, they can produce models with similar performance as individual models while being able to switch between different proposal numbers. This validates the advantage of using our switchable proposals in real applications where our model can instantly switch among different numbers of proposals when the computational resource changes.

\begin{figure}[!tb]
    \centering
    \includegraphics[width=8cm]{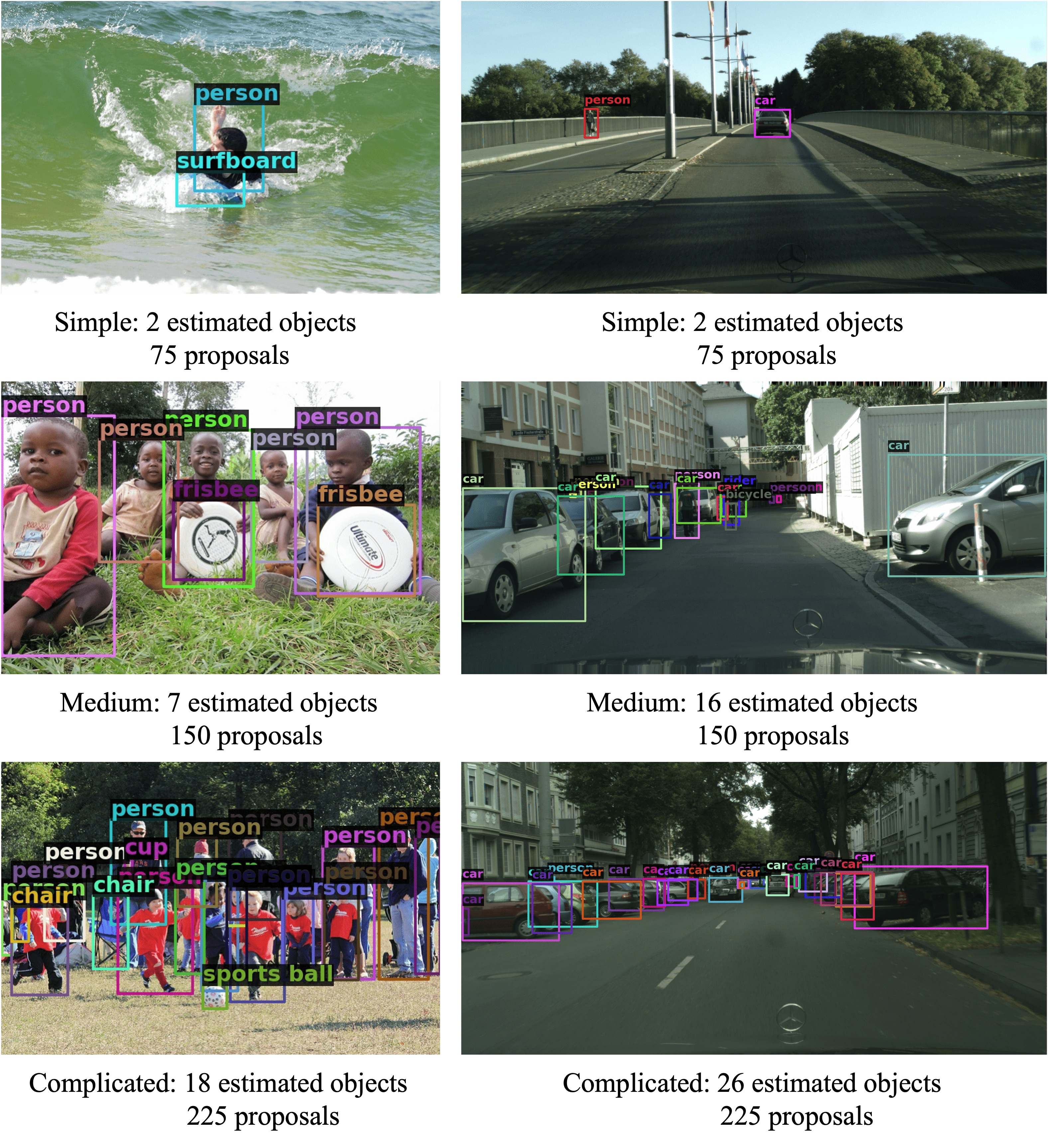}
    \caption{Object detection with dynamic proposals and their corresponding estimated numbers of objects and number of proposals. First column shows images from MS-COCO and second column shows images from Cityscapes. }
    \label{fig: proposalVisual}
    \vspace{-0.2cm}
\end{figure}

\noindent\textbf{Dynamic Proposals}
We first experiment with dynamic proposals on two-stage and query-based methods on MS COCO~\cite{lin2015microsoft}. We experiment with Cascade R-CNN, Faster R-CNN and Sparse R-CNN for object detection, and Mask R-CNN, Cascade Mask R-CNN and QueryInst~\cite{Fang_2021_ICCV} for instance segmentation. We also compare with existing methods RetinaNet~\cite{lin2017focal}, DETR~\cite{detr} and Deformable-DETR~\cite{zhu2021deformable} on object detection. As shown in Table \ref{tab: objectDetection}, integrated with our proposed dynamic proposals, both two-stage methods and query-based methods achieve similar or even better average precision (AP) while accelerating the inference by a large margin. For example, Cascade R-CNN+DP increases FPS by $16.3 \%$ compared to its original version on object detection. QueryInst+DP on ResNet50 increases FPS by $85\%$ compared to its static counterpart on instance segmentation. \textcolor{black}{The main difference in the FPS gain of object detectors is due to the different computational cost of their ROI feature processing stage. For example, Cascade Mask R-CNN has a cascaded ROI processing stage that Mask R-CNN does not have, leading to more speed up of Cascade Mask R-CNN than Mask R-CNN. QueryInst has heavier feature refinement stages than Sparse R-CNN which leads to more speedup on QueryInst. } \textcolor{black}{Note we do not include experiments of dynamic proposals on DETR and Deformable-DETR, because the numbers of proposals do not affect their inference speed significantly. }
We also conduct experiments on Cityscapes benchmark to evaluate dynamic proposals with two-stage detection methods Faster R-CNN and Mask R-CNN, and query-based methods Sparse R-CNN and QueryInst. The results are summarized in Table \ref{tab: cityscapes}. When integrated with our dynamic proposals, all of the four methods achieve similar performance on AP and AP$_\text{50}$ metrics with faster inference. For example, QueryInst + DP speeds up about $2\times$ while slightly dropping $0.8\%$ on AP. Sparse R-CNN + DP improves its speed by about $10\%$ with a slight $0.7\%$ drop on AP. Comparing the performance on Cityscapes with MS-COCO, we believe the main reason for the slight performance drop is that the smaller object scales on Cityscapes make our model prone to missing some objects when it dynamically selects fewer learned proposals.
\subsection{Model Analysis}
In this section, we conduct experiments to study the design and hyper-parameters of our proposed method. We use Sparse R-CNN \cite{sun2021sparse} and QueryInst \cite{Fang_2021_ICCV} with ResNet-50 \cite{he2015deep} as the base models to conduct experiments. 

\noindent\textbf{Study of dynamic proposals.} We visualize the final object detection results together with estimated object numbers of Sparse R-CNN \cite{sun2021sparse} on both MS COCO \cite{lin2015microsoft} and Cityscapes \cite{Cordts2016Cityscapes} in Figure~\ref{fig: proposalVisual}.  As shown in the figure, our proposed model can give a reasonable estimation of the number of objects in the images. For complicated images with multiple objects shown in the last column, more proposals are applied to guarantee detection performance. For simple images with a few objects shown in the first column, a small number of proposals are generated to increase inference speed. Based on the generated dynamic proposals, our model can efficiently localize and categorize the objects in the images with dynamic computational costs.
\begin{table}[!tb]
    \centering
    \begin{tabular}{c|c|c|c|c}
    \toprule
    \multirow{2}{*}{Method} & \multicolumn{2}{c|}{MS COCO} & \multicolumn{2}{c}{Cityscapes} \\
    \cline{2-5} 
    \rule{0pt}{10pt}
    & Diff. & Acc. & Diff. & Acc.\\ 
    \midrule
    Sparse R-CNN + DP & 1.965 & 99.4$\%$ & 4.957 & 98.5$\%$\\
    QueryInst + DP & 1.990 & 99.3$\%$ & 4.657 & 98.1$\%$\\
    Faster R-CNN + DP & 2.135 & 98.9$\%$ & 4.997 & 97.9$\%$\\
    Mask R-CNN + DP & 2.251 & 98.6$\%$ & 4.891 & 97.5$\%$\\
    \bottomrule
    \end{tabular}
    \caption{Average differences between our estimated numbers of objects and ground truths and the corresponding 4-class classification accuracies on different benchmarks with dynamic proposals.}
    \label{tab: est_obj}
    \vspace{-0.2cm}
\end{table}
Meanwhile, we report the average differences between our estimated numbers of objects and ground truths on both MS COCO \texttt{val} and Cityscapes \texttt{val} benchmarks as in Table \ref{tab: est_obj}. We also report the accuracies of the estimated number of objects by considering the process of assigning an input image to the four choices of proposals as a classification task. From the table, the mean errors between our estimated numbers of objects and ground truths are within $5$ objects on both MS COCO and Cityscapes. Under the 4-class classification metric, the four experimented methods can all achieve accuracies higher than $97\%$ on both MS COCO and Cityscapes, proving the effectiveness of the simple regression network $\mathcal{M}$. 
We also conduct experiments using $n_g$ as $\tilde{n}$ in Equation \ref{eq: delta} during inference as an oracle result on Sparse R-CNN + DP, shown in Table \ref{tab: est_obj_vs_gt}. When using ground truth $n_g$, the APs on MS COCO and Cityscapes are very close to those with estimated $\tilde{n}$, which indicates the estimated number of objects is already accurate enough to select the number of proposals.
\begin{table}[!tb]
    \centering
    \begin{tabular}{c|c|c|c|c}
    \toprule
    \multirow{2}{*}{Method} & \multicolumn{2}{c|}{MS COCO} & \multicolumn{2}{c}{Cityscapes}\\
    \cline{2-5} 
    \rule{0pt}{10pt}
    & AP & AR & AP & AP$_{50}$\\
    \midrule
    Ground truth $n_g$ & 45.2 & 65.4 & 41.3 & 66.6\\
    Estimated $\tilde{n}$ & 45.3 & 65.4 & 41.1 & 66.4\\
    \bottomrule
    \end{tabular}
    \caption{Comparison between using ground truth $n_g$ and estimated numbers of objects $\tilde{n}$ in Equation  \ref{eq: delta} during inference on Sparse R-CNN + DP.}
    \label{tab: est_obj_vs_gt}
    \vspace{-0.2cm}
\end{table}

\noindent\textbf{Study of inplace distillation.} We conduct experiments to evaluate the effectiveness of inplace distillation on QueryInst + DP and the results are summarized in Table \ref{tab: distillation}. It can be seen that by using inplace distillation, AP of QueryInst + DP can be improved by at least $0.3\%$ while keeping similar AR results for both object detection and instance segmentation. This proves the effectiveness of our feature-level inplace distillation which helps model variants with few proposals to learn from the full model with most proposals \textcolor{black}{ and stabilize training of our models}. 
\begin{table}[!tb]
    \centering
    \begin{tabular}{c|c|c|c|c}
    \toprule
   Inplace & \multicolumn{2}{c|}{Obj. Det.} & \multicolumn{2}{c}{Ins. Seg.}\\
    \cline{2-5} 
    \rule{0pt}{10pt}
    Distillation & AP & AR & AP & AR \\
    \midrule
    $\times$ & 44.7 & 65.5 & 41.4 & 59.5 \\
    $\checkmark$ & 45.3 & 65.4 & 41.7 & 59.6\\
    \bottomrule
    \end{tabular}
    \caption{Performance comparison of QueryInst + DP with and without inplace distillation.}
    \label{tab: distillation}
    \vspace{-0.2cm}
\end{table}

\noindent\textbf{Study of configuration numbers $\theta$.} To investigate the effect of configuration number $\theta$, we conduct experiments on QueryInst + DP with $\theta = 4, 8$ and $12$, respectively. Results are shown in Table \ref{tab:configurations}. It is shown that different configurations do not impact speed of the model notably.
Comparing model $A$ and $B$, by increasing the number of configurations, the performance of both object detection and instance segmentation are improved by around $0.2\%$ across all metrics. Comparing model $B$ and $C$, the performances on both AP and AR tend to be saturated.
\begin{table}[!tb]
    \centering
    \begin{tabular}{c|c|c|c|c|c}
    \toprule
    Model & AP-Det & AR-Det & AP-Seg & AR-Seg & FPS \\
    \midrule
    $\theta=4$ & 47.3 & 67.9 & 41.7 & 59.6 & 7.4\\
    $\theta=8$ & 47.5 & 68.2 & 41.9 & 59.8 & 7.3 \\
    $\theta=12$ & 47.5 & 68.2 & 41.7 & 60.2 & 7.3\\
    \bottomrule
    \end{tabular}
    \caption{Comparison of instance segmentation results with different configuration numbers.}
    \label{tab:configurations}
    \vspace{-0.2cm}
\end{table}

\section{Conclusion}
In this paper, we present a framework to dynamically adjust the number of proposals for two-stage and query-based detection methods based on input images or current computational resources. Equipped with switchable proposals, we find that a single query-based model is able to switch to different numbers of proposals while achieving similar performance as individual models. Equipped with dynamic proposals, one model can adaptively choose a subset of proposals according to the estimated complexity of the input image, reducing the average inference time. We hope our work can inspire the community to the development of dynamic and efficient deep models.

{\small
\bibliographystyle{unsrt}
\bibliography{egbib}
}
\end{document}


\title{Supplementary of Dynamic Proposals for Efficient Object Detection}

\maketitle
\section{Main Results}
\begin{figure} [!tb]
    \centering
    \includegraphics[width=8cm]{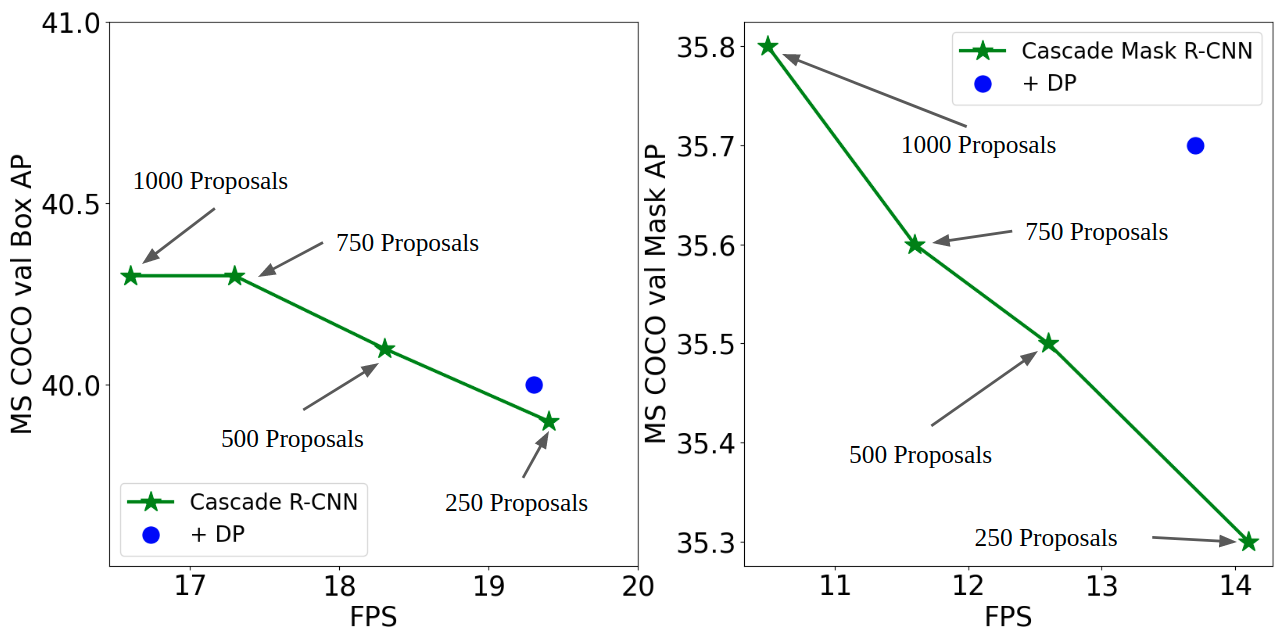}
    \caption{AP $vs.$ FPS on MS COCO \texttt{val} benchmarks. Integrated with our dynamic proposals, the inference speeds of the four shown detection methods increase by a large margin while maintaining competitive performance. The inference speed is measured with a single TITAN RTX GPU.}
    \label{fig:cascade}
\end{figure}

\noindent\textbf{Experiments on more models on switchable proposals.} We provide experiments on more two-stage models integrated with switchable proposals as Table \ref{tab: moreSwitch} and Figure \ref{fig:cascade}.  

\begin{table*}[!tb]
    \centering
    \begin{tabular}{c|c|c|c|c|c|c|c|c|c|c}
    \toprule
     \multirow{2}{*}{Model} & \multirow{2}{*}{Method} & \multicolumn{2}{c|}{0.25N} & \multicolumn{2}{c|}{0.5N} & \multicolumn{2}{c|}{0.75N} & \multicolumn{2}{c|}{N} & Train \\
     \cline{3-10} \rule{0pt}{10pt}
     & & AP & AR & AR & AR & AP & AR & AP & AR & Time\\
    \midrule
    \multirow{4}{*}{Cascade R-CNN} & Individual & 39.8 & 52.3 & 40.1 & 52.5 & 40.3 & 52.9 &  40.3 & 53.0 & $\sim$ 50\\
    & Switchable & 39.9 & 52.1 & 40.3 & 52.3 & 40.3 & 52.8 & 40.3 & 53.2 & $\sim$ 16\\
    & Naive Selection & 30.4 & 40.6 & 36.5 & 49.1 & 38.9 & 52.3 & 40.3 & 52.9 & - \\
    \cline{3-10} \rule{0pt}{10pt}
    & FPS & \multicolumn{2}{c|}{19.4}& \multicolumn{2}{c|}{18.3}& \multicolumn{2}{c|}{17.3} & \multicolumn{2}{c|}{16.6} & -\\
    \midrule
    \multirow{4}{*}{Cascade Mask R-CNN} & Individual & 35.6 & 47.0 & 35.8 & 47.8 & 35.8 & 48.1 & 35.9 & 48.3 & $\sim$ 70\\
    & Switchable & 35.3 & 46.4 & 35.5 & 46.7 & 35.6 & 47.1 & 35.8 & 47.6 & $\sim$ 20 \\
    & Naive Selection & 28.2 & 37.6 & 33.2 & 44.7 & 35.0 & 46.9 & 35.9 & 48.3 & - \\
    \cline{3-10} \rule{0pt}{10pt}
    & FPS & \multicolumn{2}{c|}{14.1}& \multicolumn{2}{c|}{12.8}& \multicolumn{2}{c|}{11.6} & \multicolumn{2}{c|}{10.5} & -\\
    \bottomrule
    \end{tabular}
    \caption{More examples of comparisons between original two-stage models trained with multiple configurations individually and jointly trained when integrated with switchable proposals on MS COCO \texttt{val} benchmark. Training time is estimated in GPU hours.}
    \label{tab: moreSwitch}
\end{table*}

\section{Model Analysis}
\noindent\textbf{Study of configuration numbers $\theta$ on switchable proposals.} We conduct experiments on switchable proposals with different numbers of configurations with both Sparse R-CNN and QueryInst on MS COCO benchmarks. We train $8$ baseline models with different numbers of proposals, namely $40, 75, 115, 150, 190$, $225, 265$ and $300$, with the original Sparse R-CNN and QueryInst individually and use their corresponding number of proposals during the inference process, denoted as ``Individual". For models with switchable proposals, we use 300 proposals as $N_{s}$ during the training process. During the inference process, we assign $N_{s}$ with different numbers, namely $40, 75, 115, 150, 190$, $225, 265$ and $300$ to get $8$ inference results, which is denoted as ``Switchable". Note the switchable results are obtained from a single model with switchable proposals. Finally, we select the Sparse R-CNN model with $300$ proposals and naively select the first $40, 75, 115, 150, 190, 225, 265$, and $300$ elements from the $300$ proposals during inference to produce $8$ inference results, denoted as ``Naive selection". Table \ref{tab: switchObj} and \ref{tab: switchIns} summarize the comparison among these trained models with different numbers of proposals during inference processes. 

From Table \ref{tab: switchObj} and \ref{tab: switchIns}, the performances on AP and AR of separately and jointly training are quite similar given the same number of proposals. The FPS of ``switchable'' and ``individual'' models under the same number of proposals are the same since they have exactly the same computation procedure. Our ``switchable" model only needs at most about $40\%$ training time of the ``individual" approach, which needs to train $8$ models separately. For the ``naive selection'' setup, the performance drops a lot compared with the other two models, especially when the number of proposals is reduced to 150 or even smaller. This experiment shows that by naively selecting the first $N_{s}$ proposals of the original query-based methods, the performance degenerates significantly. Therefore, the original models are not suitable to be directly used for multiple configurations by simply sampling the original proposals during the inference process. However, by using our switchable proposals to train query-based methods, the training time can be saved considerably while keeping similar performance on both efficiency and accuracy. This validates the advantage of using our switchable proposals in real applications where our model can instantly switch among different numbers of proposals when the computational resource changes.
\begin{table}[!tb]
    \centering
    \begin{tabular}{c|c|c|c|c|c|c|c}
     \toprule
    \multirow{2}{*}{Prop.} & \multicolumn{2}{c|}{Individual} & \multicolumn{2}{c|}{Switchable} & \multicolumn{2}{c|}{Naive} & \multirow{2}{*}{FPS}\\
    \cline{2-7} \rule{0pt}{10pt}
    & AP & AR & AP & AR & AP & AR & \\
    \midrule
    40 & 35.3 & 46.9 & 35.5 & 48.7 & 14.4 & 29.6 & 19.9\\
    75 & 42.6 & 58.3 & 42.6 & 59.7 & 26.5 & 47.4 & 19.5\\
    115 & 44.1 & 60.5 & 44.1 & 61.5 & 33.6 & 53.5 & 19.2\\
    150 & 44.2 & 63.2 & 44.3 & 64.4 & 37.2 & 58.0 & 18.8\\
    190 & 44.5 & 63.7& 44.5 & 65.2 & 40.8 & 60.6 & 18.5\\
    225 & 44.6 & 64.2 & 44.7 & 66.0 & 43.5 & 64.4 & 18.1\\
    265 & 44.8 & 65.7 & 44.8 & 66.3 & 44.6 & 65.6 & 17.9\\
    300 & 45.0 & 66.7 & 45.3 & 66.7 & 45.0 & 66.7 & 17.7\\
    \midrule
    Train & \multicolumn{2}{c|}{$\sim$ 200} & \multicolumn{2}{c|}{$\sim$ 80} & \multicolumn{2}{c|}{-} & -\\
    \bottomrule
    \end{tabular}
    \caption{Comparison between original Sparse R-CNN trained with $8$ configurations individually and jointly with switchable proposals on MS COCO \texttt{val} benchmark. ``Train" denotes the training time estimated in GPU hours.  Inference speed is tested on a single TITAN RTX GPU.}
    \label{tab: switchObj}
\end{table}

\begin{table}[!bt]
    \centering
    \begin{tabular}{c|c|c|c|c|c|c|c}
     \toprule
    \multirow{2}{*}{Prop.} & \multicolumn{2}{c|}{Individual} & \multicolumn{2}{c|}{Switchable} & \multicolumn{2}{c|}{Naive} & \multirow{2}{*}{FPS}\\
    \cline{2-7} \rule{0pt}{10pt}
    & AP & AR & AP & AR & AP & AR & \\
    \midrule
    40 & 35.2 & 47.1 & 35.1 & 46.8 & 27.0 & 40.6 & 12.5\\
    75 & 39.4 & 53.5 & 39.4 & 53.2 & 28.2 & 47.5 & 9.4\\
    115 & 39.8 & 55.8 & 39.7 & 55.7 & 37.8 & 53.4 & 7.5\\
    150 & 40.8 & 57.2 & 40.9 & 57.3 & 37.7 & 55.6 & 6.4\\
    190 & 41.3 & 58.2& 41.1 & 58.4 & 40.3 & 57.6 & 5.4\\
    225 & 41.5 & 58.8 & 41.1 & 59.0 & 41.0 & 58.9 & 4.7\\
    265 & 41.6 & 59.5 & 41.4 & 59.9 & 41.5 & 59.7 & 4.2\\
    300 & 41.7 & 59.9 & 41.4 & 60.0 & 41.7 & 59.9 & 3.7\\
    \midrule
    Train & \multicolumn{2}{c|}{$\sim$370} & \multicolumn{2}{c|}{$\sim$100} & \multicolumn{2}{c|}{-} & -\\
    \bottomrule
    \end{tabular}
    \caption{Comparison between original QueryInst trained with $8$ configurations individually and jointly with switchable proposals on MS COCO \texttt{val} benchmark.  ``Train" denotes the training time estimated in GPU hours. Inference speed is tested on a single TITAN RTX GPU.}
    \label{tab: switchIns}
\end{table}

\noindent\textbf{Study of configuration numbers $\theta$ on dynamic proposals.} To investigate the effect of configuration number $\theta$, we conduct experiments on Sparse R-CNN + DP with $\theta = 4, 8$ and $12$, corresponding to models $A$, $B$ and $C$, respectively. Results are shown in Table \ref{tab:configurations}. It is shown that different configurations do not impact speed of the model notably. Comparing model $A$ and $B$, by increasing the number of configurations, the performance of AP and AR drops a little bit for most of the metrics besides those for small objects. Comparing model $B$ and $C$, the performances on both AP and AR tend to be saturated.
\begin{table*}[!bt]
    \centering
    \begin{tabular}{c|c|c|c|c|c|c|c|c|c}
    \toprule
     Model & AP & AP(S) & AP(M) & AP(L) & AR & AR(S) & AR(M) & AR(L) & FPS \\
    \midrule
    A & 45.3 & 27.5 & 48.4 & 59.8 & 65.4 & 45.3 & 69.9 & 81.9 & 19.0 \\
    B & 45.1 & 27.7 & 48.1 & 59.1 & 65.6 & 45.5 & 69.2 & 81.1 & 19.0 \\
    C & 45.0 & 27.8 & 48.5 & 58.9 & 65.0 & 45.5 & 68.7 & 81.6 & 18.8 \\
    \bottomrule
    \end{tabular}
    \caption{Comparison of object detection results with different configuration numbers on MS COCO \texttt{val}. Inference speed is tested on a single TITAN RTX GPU.}
    \label{tab:configurations}
\end{table*}

\begin{figure} [!t]
    \centering
    \includegraphics[width=8cm]{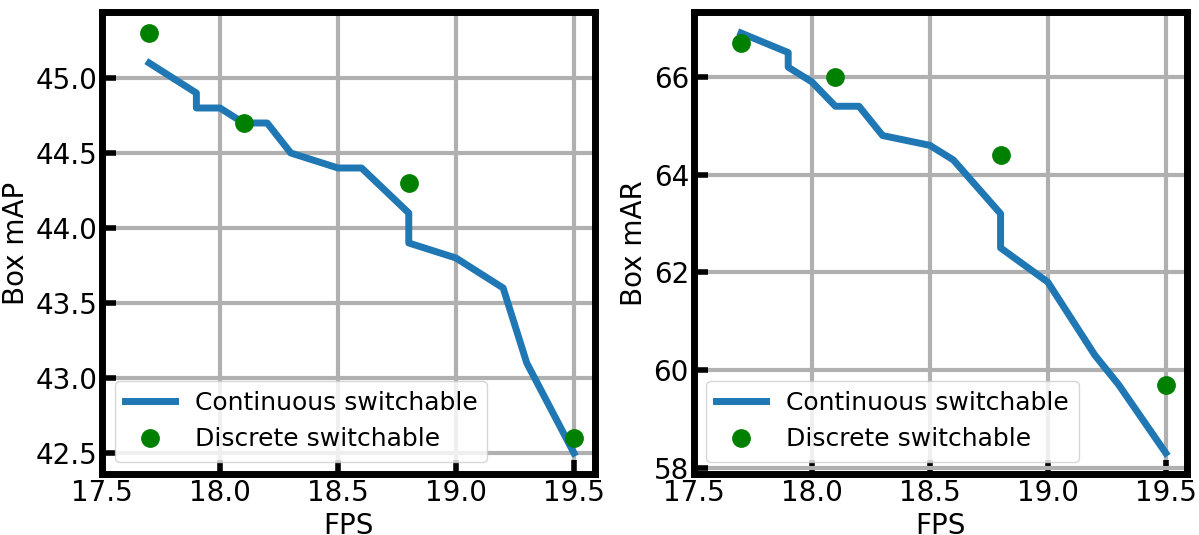}
    \caption{Sparse R-CNN with discrete / continuous numbers of proposals.}
    \label{fig:multipleProp}
\end{figure}

\textcolor{black}{To verify whether our approach work with arbitrary selection of proposal numbers, we also conduct experiments on Sparse R-CNN with a continuous list of proposal numbers ranging from 75 to 300 in the switchable setting. The model is then evaluated exhaustively with proposal numbers from 75 to 300 with an interval of 5. Figure \ref{fig:multipleProp} shows that the performance of the continuous selection is comparable to the original discrete selection, which indicates our model is compatible with continuous selection of proposal numbers.}

\textcolor{black}{We also conduct experiments to use a 300-proposal detector to distill a 75-proposal detector and compare with our proposed method in Table \ref{tab: distill}. The distillation-only method is worse than our switchable method.}
\begin{table}[!tb]
    \centering
    \begin{tabular}{c|c|c}
    \toprule
    Method & AP & AR\\
    \midrule
    Sparse R-CNN + Distill & 42.2 & 58.3\\
    Sparse R-CNN + SwitchProp & 42.6 & 59.7\\
    \bottomrule
    \end{tabular}
    \caption{Comparison between using a 300-proposal detector to distill a 75-proposal detector and the proposed method.}
    \label{tab: distill}
\end{table}

\noindent\textbf{Study of sampling function.} For two-stage object detectors, the proposals after RPN are sorted based on their confidence scores. Therefore, the first $N_s/N_d$ elements are always proposals with higher confidences than those thrown. For query-based methods, there is no specific order for the proposals. We conduct experiments with dynamic proposals on the choices of sampling as Table \ref{tab:sampling}. ``First'' and ``Last'' mean choosing the first and last $N_d$ proposals while ``Bin'' represents binning 300 proposals into 75 buckets and sample $1,2,3,4$ proposals from each bucket. For two-stage methods, the results of sampling the first $N_d$ proposals are the best while those of sampling the last are the worst. For query-based methods, since the proposals do not have specific order, sampling strategy has little influence on the accuracy as long as the same sampling strategies are applied to training and inference.

\begin{table}[!bt]
    \centering
    \begin{tabular}{c|c|c|c|c|c|c}
    \toprule
    & \multicolumn{3}{c|}{QueryInst} & \multicolumn{3}{c}{Faster R-CNN}\\
    \cline{2-7}
    & Bin & First & Last & Bin & First & Last\\
    \midrule
    AP & 41.6 & 41.7 & 41.6 & 36.9 & 37.5 & 12.1\\
    AP$_s$ & 21.9 & 22.6 & 22.3 & 20.0 & 20.2 & 6.7\\
    AP$_m$ & 44.2 & 44.8 & 44.7 & 40.0 & 40.2 & 17.7\\
    AP$_l$ & 59.8 & 59.9 & 59.7 & 49.3 & 49.8& 12.1 \\
    \bottomrule
    \end{tabular}
    \caption{Comparison of AP on different sampling strategies.}
    \label{tab:sampling}
\end{table}

\noindent\textbf{Analysis of inference time.} We provide the time ratio of proposal processing of the evaluated models with ResNet50 as in Table \ref{tab:inference}. Methods with multiple stages of computation post proposals like QueryInst and Cascade R-CNN gain more speedup because their task heads account for a larger ratio in computation.

\begin{table}[!bt]
    \centering
    \begin{tabular}{c|c}
    \toprule
    Methods & Inference time ratios\\
    \midrule
     QueryInst & 75.22$\%$\\
     Cascade R-CNN & 36.24$\%$ \\
     Faster R-CNN & 13.43$\%$\\
     Mask R-CNN & 21.38$\%$\\
    \bottomrule
    \end{tabular}
    \caption{Inference time ratios of task heads}
    \label{tab:inference}
\end{table}

\section{More visualization results.}
\begin{figure*}[!bt]
    \centering
    \includegraphics[width=15cm]{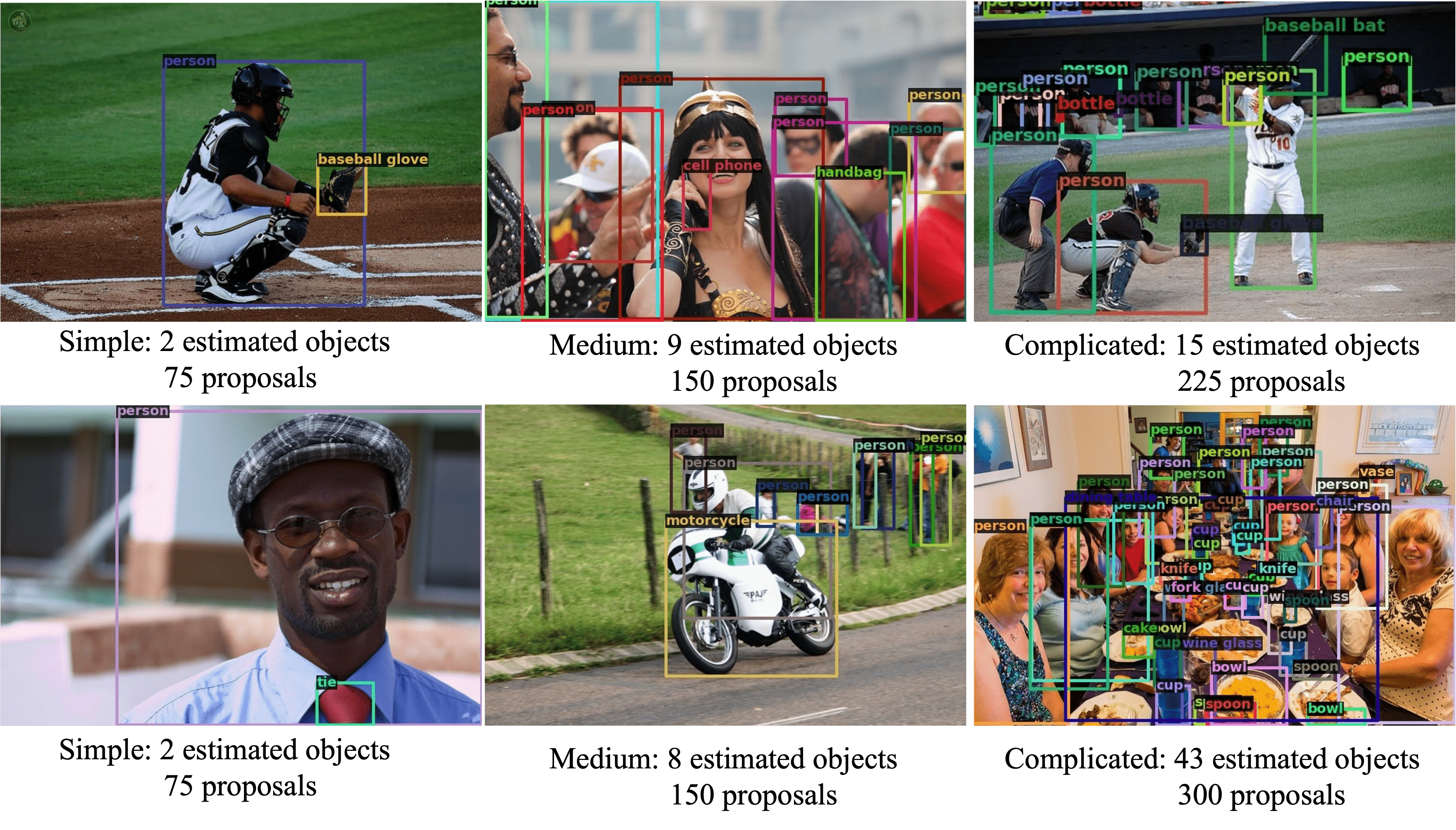}
    \includegraphics[width=15cm]{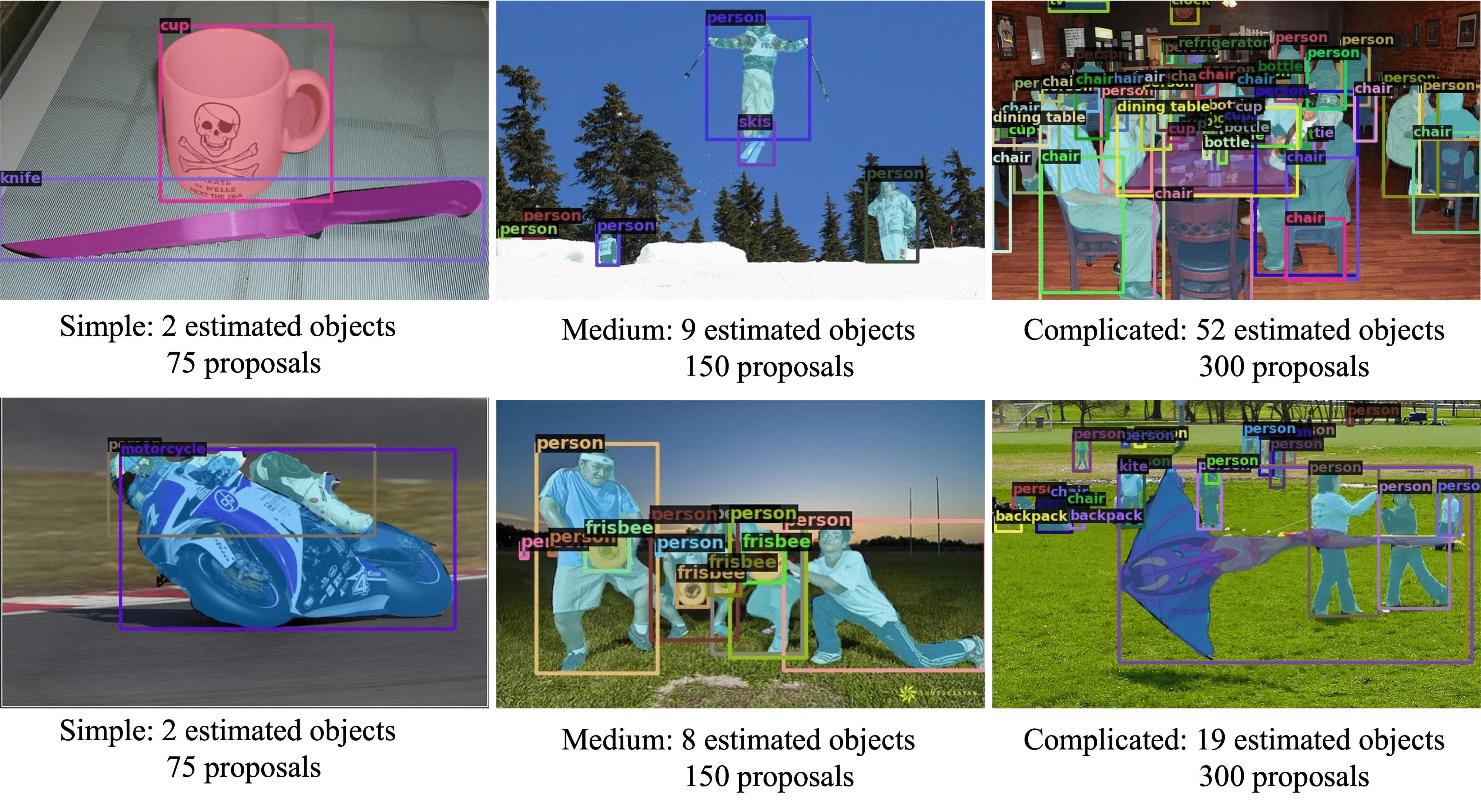}
    \caption{Object detection with dynamic proposals and their corresponding estimated numbers of objects and number of proposals on MS COCO benchmark. First two rows show results of object detection with Sparse R-CNN + DP and last two rows show results of instance segmentation with QueryInst + DP. }
    \label{fig: visualEx}
\end{figure*}

\begin{figure*}[!bt]
    \centering
    \includegraphics[width=16cm]{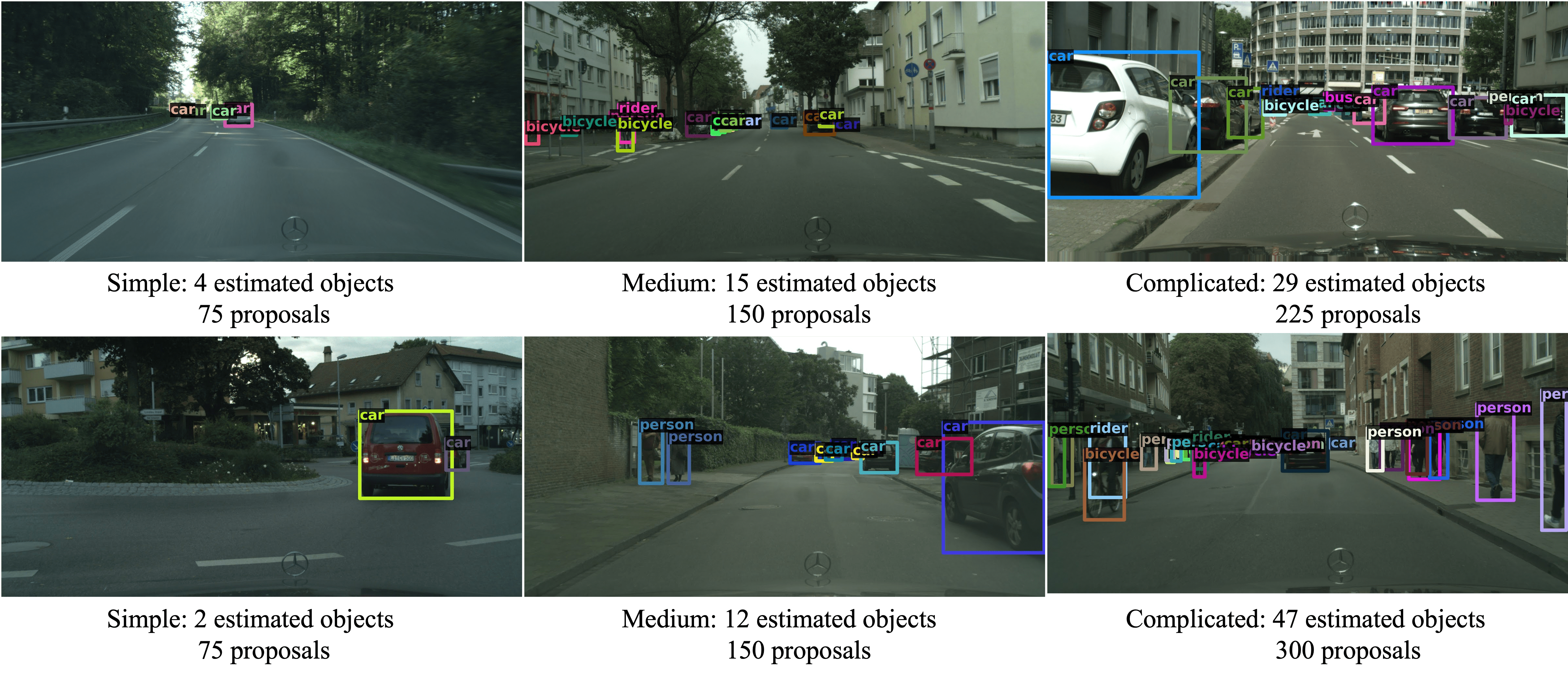}
    \includegraphics[width=16cm]{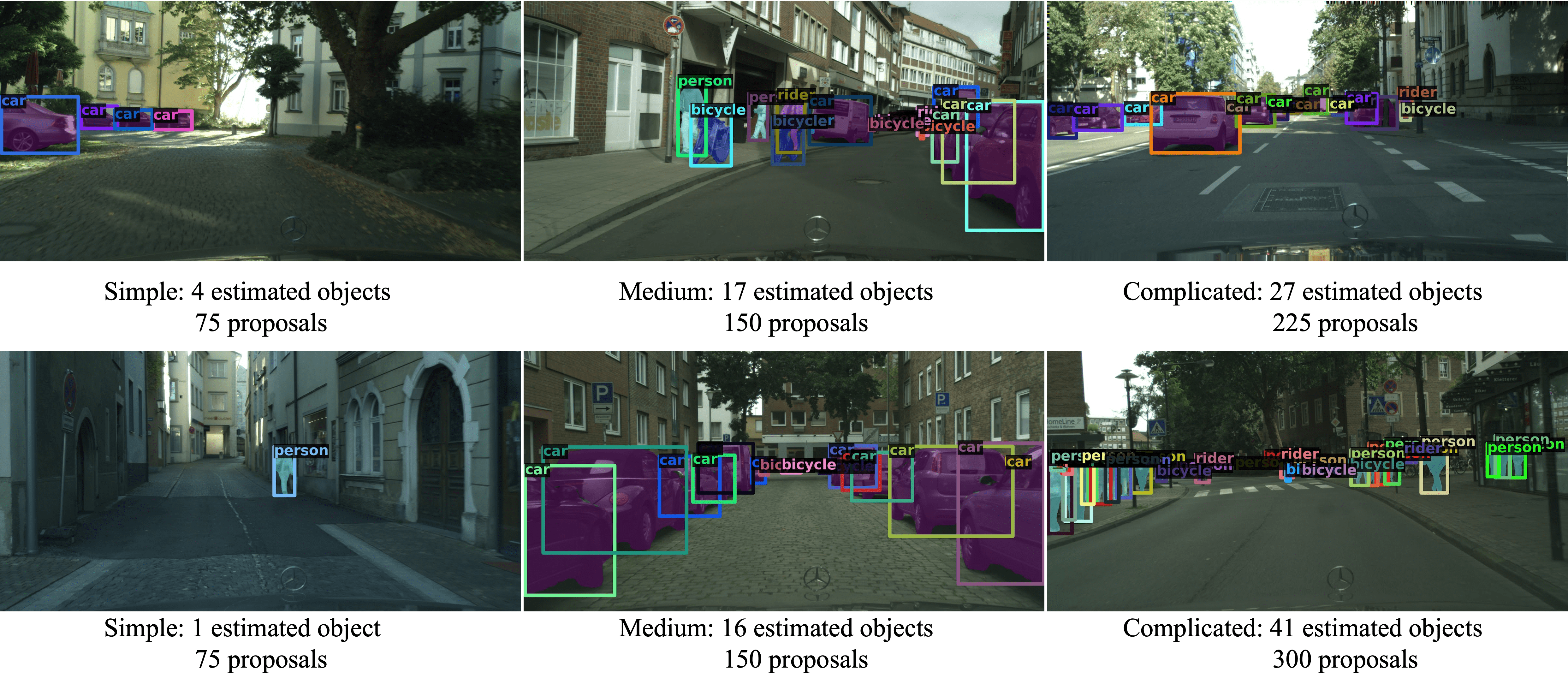}
    \caption{Object detection with dynamic proposals and their corresponding estimated numbers of objects and number of proposals on Cityscapes benchmark. First two rows show results of object detection with Sparse R-CNN + DP and last two rows show results of instance segmentation with QueryInst + DP. }
    \label{fig: visualEx2}
\end{figure*}

We also provide more visualization results of object detection with Sparse R-CNN + DP and instance segmentation with QueryInst + DP on MS COCO and Cityscapes benchmarks as Figures \ref{fig: visualEx} and \ref{fig: visualEx2}.

{\small
\bibliographystyle{ieee_fullname}
\bibliography{supp}
}